\documentclass[letterpaper, 10 pt, conference]{ieeeconf}  

\IEEEoverridecommandlockouts                              
\usepackage{graphics} 
\usepackage{epsfig} 
\usepackage{amsmath} 
\usepackage{amssymb}  

\usepackage{url}
\usepackage[ruled, vlined, linesnumbered]{algorithm2e}
\usepackage{verbatim}
\usepackage{soul, color}
\usepackage{lmodern}
\usepackage{fancyhdr}
\usepackage[utf8]{inputenc}
\usepackage{fourier}
\usepackage{array}
\usepackage{makecell}

\SetNlSty{large}{}{:}

\makeatletter

\newcommand{\Rom}[1]{\expandafter\@slowromancap\romannumeral #1@}
\makeatother

\title{\LARGE \bf
Light-weighted Saliency Detection with Distinctively Lower Memory Cost and Model Size
}

\author{Shanghua Xiao
\\ Department of Computer Science, Sichuan University, Sichuan Province, China \\
{\tt 2015223040036@stu.scu.edu.cn}
}

\begin{document}

\maketitle
\thispagestyle{plain}
\pagestyle{plain}

\begin{abstract}
Deep neural networks (DNNs) based saliency detection approaches have succeed in recent years, and improved the performance by a great margin via increasingly  sophisticated network architecture. Despite the performance improvement, the computational cost is excessively high for such low level visual task. In this work, we propose a light-weighted saliency detection approach with distinctively lower runtime memory cost and model size. We evaluated the performance of our approach on multiple benchmark datasets, and achieved competitive results comparing with state-of-the-art methods on multiple metrics. We also evaluated the computational cost of our approach with multiple measurements. The runtime memory cost of our approach is 42 to 99 times fewer comparing with the previous DNNs based methods. The model size of our approach is 63 to 129 times smaller, and takes less than 1 Megabytes storage space with out any deep compression technique.
\end{abstract}

\section{INTRODUCTION}
\par Saliency detection aims to extract a mask map, termed "saliency map", that gives the probability of most attractive part from a given image. Such a saliency map can be utilized for multiple applications, such as image thumbnailing \cite{b1} and content-aware image resizing \cite{b2}. By reducing the scale of the field-to-be-perceived to specific salient region, it can also be used as preprocessing method to further speed-up other visual tasks, e.g. \cite{b3,b4}. The extensive usages leads to the popularity of saliency detection, and recently many approaches are proposed towards this issue.

\par Recent approaches adopt deep convolutional neural networks (DCNNs) to boost up saliency detection. Multiple DCNNs models such as \cite{b11,b20} have revolutionized the area of large scale image classification, and then extensively used in multiple visual tasks with transfer learning, such as target tracking, image segmentation and also saliency detection. Previous DCNNs based saliency detection approaches, for example, DeepFix \cite{b6} utilize VGG16 to initiate the parameter of its lower layers, and DeepGazeII \cite{b18} utilize VGG19 to initiate its whole feature extraction layers. Though the representative power of transferred models help previous methods achieve high performance, the computational cost also raises as the transferred models become increasingly computational and storage intensive.

\begin{figure}[!t]
\centering
    \includegraphics[width=3.3in]{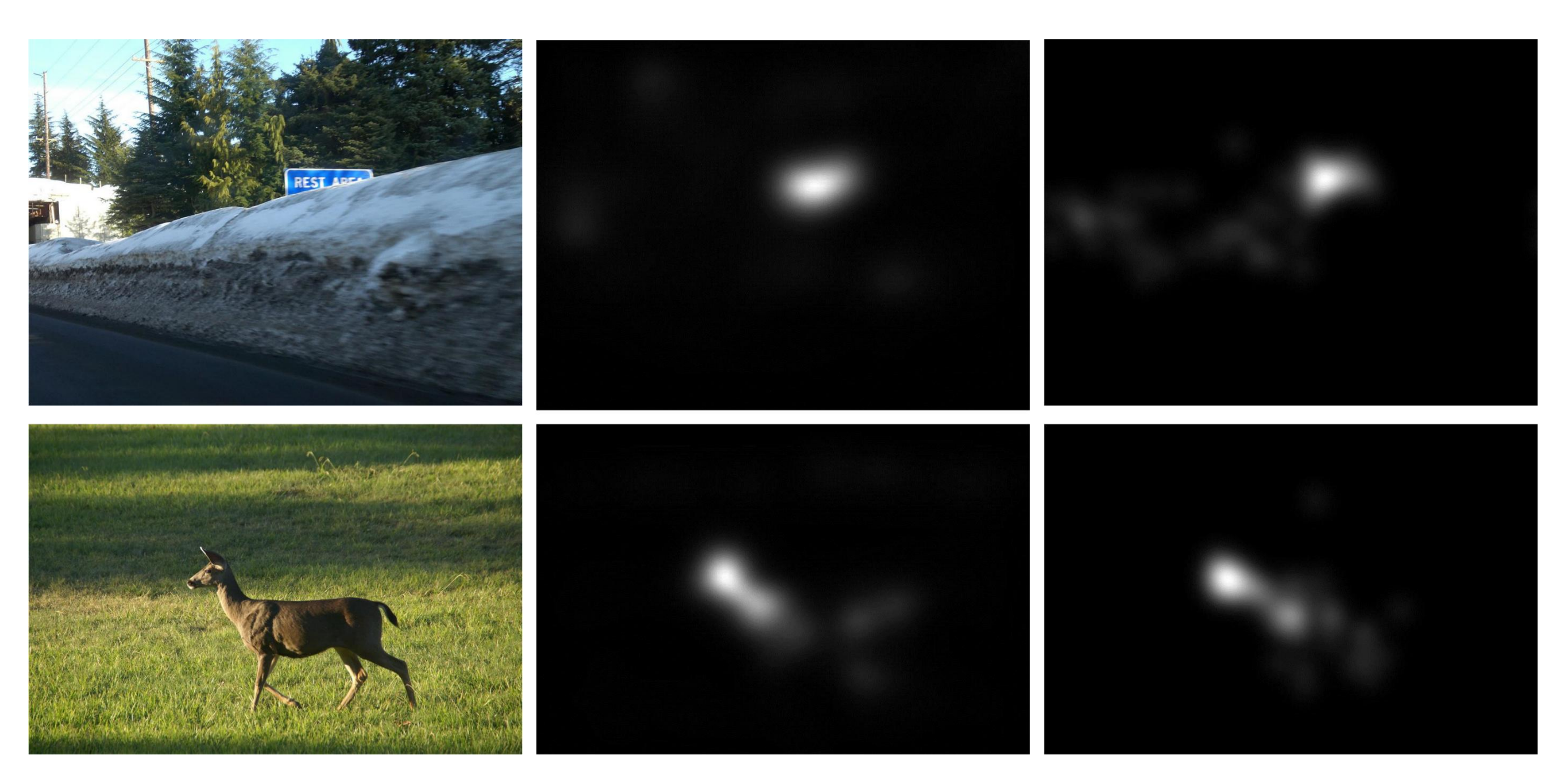}
    {Comparison between ground truth and output saliency map of our approach.\label{fig1}}
\end{figure}

\par The computational cost of DCNNs can be mainly described as the network workload, which is the theoretical number of basic operations needed in the DCNNs computation from the algorithmic aspect \cite{b51}. It can also present as the temporary intermediate data size which generated from the forwarding layer-wise calculation, and the model size determined by the parameter number. From hardware perspective, the computational cost can be described by CPU, RAM and disk consumption. Recently, DNNs deployment on mobile device receives increasing attention, while the computational power of mobile devices is limited comparing with PC and server. Since the gap between high computational complexity of DNNs and low computational capacity of mobile devices exists, multiple methods were proposed towards this issue. SqueezeNet \cite{b52} proposed a bottleneck structure to reduce the computational cost by reducing the channels before feature extraction. MobileNet \cite{b50} introduced the inverted residual block composed with depthwise separable convolution, and achieved AlexNet \cite{b17} level accuracy with much lower computational cost.

\begin{figure*}
\centering
    \includegraphics[width=7.0in]{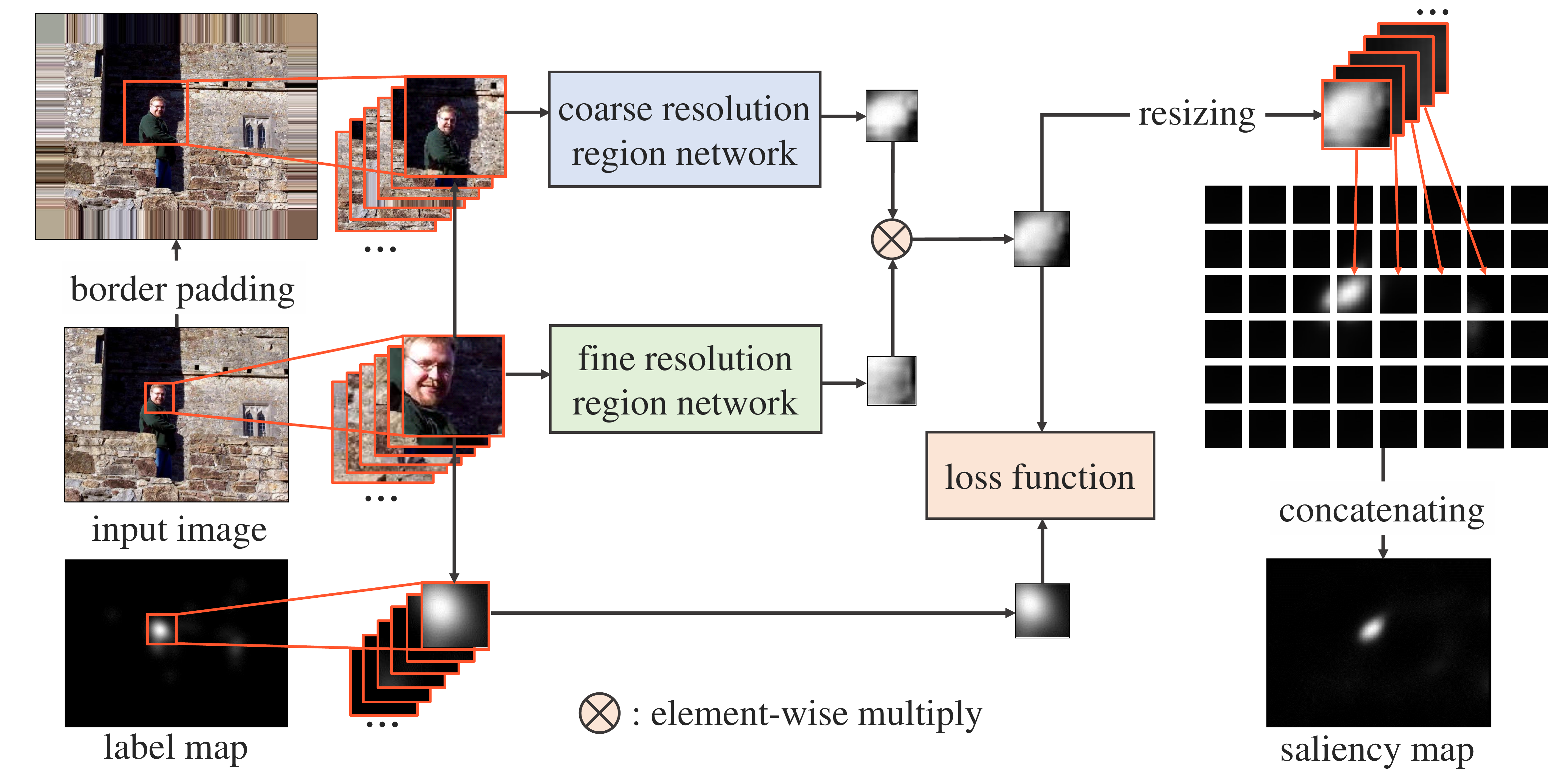}
    {The pipeline of our approach.\label{fig1}}
\end{figure*}

\par As a simulation for early stage visual attention task, saliency detection naturally requires for low computational cost, yet few effort have been made toward this direction. In this work, we propose a novel saliency detection approach with distinctively lower computational cost comparing with previous methods. To achieve light-weighted network structuring, we utilize depthwise separable convolutional block to reduce the network parameter number, and we replace the processing of whole image with serial of regions to reduce the spatial scale of intermediate feature maps. We also build our approach from scratch with simpler model structure instead of transferring from trained image classification networks.

\par We first evaluate the saliency detection performance of our approach on multiple benchmark datasets, and achieve competitive results comparing with state-of-the-art methods. We then evaluate the computational cost by recording the runtime RAM cost, time consumption for processing single image and model size, and yielding distinctively lighter weighted model result.

\section{Relative Works}

\par In this section, we introduce the related works with efforts on improving saliency detection performance, and on achieving light-weighted neural network designing.

\par Traditional saliency detection approaches are mainly driven by biological and psychological studies on human attention mechanism. Following the early study "Feature Integration Theory" \cite{b8} on human visual attention mechanism, Koch \emph{et al}. propose a biological plausible visual attention model \cite{b9} which combines low level visual features such as color and contrast to produce a saliency indicating map. Later, Itti et al propose a saliency detection approach based on the behaviour and neural architecture of primate visual system \cite{b10}. They extract low level visual cues into multiple conspicuity map, and use a dynamic neural network to integrate them into a single saliency map.

\par More recently, DCNNs based approaches have improved the state-of-the-art of saliency detection by a great margin. An early work of applying CNNs to saliency detection is ensembles of Deep Networks (eDN) \cite{b15}, which models a 3 layers convolution for feature extraction and a following SVM for salient classification. Later, DeepGaze \cite{b16} adopt a deeper architecture with convolutional layers from AlexNet to extract feature maps from different levels. The feature maps then integrated into saliency map by a learned linear model. As DeepGaze introduces transfer learning into saliency detection, later approaches extensively utilize trained models from image classification to boost up the saliency detection performance. Kruthiventi et al. propose an fully convolutional neural network model named DeepFix \cite{b6}, which utilize inception module and kernel with hole to extract multiple scale features, and applies VGG-16 \cite{b11} pretrained model to initialize its early feature extraction layers. In the work of SALICON \cite{b19}, Huang et al. explored multiple pretrained models from image classification for feature extraction in saliency detection, including AlexNet, VGG-16 and GoogLeNet \cite{b20}. As the performance improves, the computational cost for DCNNs based saliency detection approaches is continuously increasing.

\par As the popularity of mobile and embedded hardware based DNNs deployment grows, the gap between limited computational capacity hardware and computational complexity of popular DNNs also enlarges. Deep Compression \cite{b53} proposed multiple techniques, namely pruning, trained quantization and Huffman coding to reduce the storage requirement of DNNs. Though Deep Compression successfully compressed the size of network model file, the runtime memory cost is not reduced since the original network structure need to be recovered from the compressed network data when performing forward computing.

\par Recently, efforts on designing efficient models from the bottom draws more attention. SqueezeNet \cite{b52} propose a novel paradigm for designing more light-weighted neural network, including replacing $3\times3$ kernels with $1\times1$ kernels and decreasing the number of input channels to $3\times3$ kernels. MobileNet \cite{b49} adopt depth-wise separable convolution to design a highly light-weighted neural network feasible for mobile device. ShuffleNet \cite{b47} utilize group convolution and channel shuffle to reduce the model size while maintaining high performance.

\par In this work, by utilizing these efficient model designing paradigm, we propose a DCNNs based saliency detection approach with distinctively lower computational cost comparing with previous methods.

\section{Proposed Approach}

\par In this section, we introduce the details of our proposed light-weighted saliency detection approach. The main objective of our work is to develop a light-weighted saliency detection approach. To achieve this, we adopt three strategies when designing the network architecture:
\begin{itemize}
\item Depthwise Separable Convolution: Replacing the normal $3\times3$ convolution layers with depth-wise separable convolutional blocks to reduce the model size from the bottom;
\item Regional Input: Processing with serial of two multi-resolution image regions instead of the whole image to reduce the feature map data blob size;
\item Simplified Networks: Constructing network with less depth and width from scratch instead of transferring from large scale models for image classification to reduce model size. We use two fully convolutional neural networks with independent parameters for feature extraction.
\end{itemize}

\par The pipeline of our approach is visualized in Fig.3, and can be briefly described as follows: 1) Raw input image and border padded input image are cropped into serial of regions at the same central location and resized into the same size; 2) The two regions are feed into two fully convolutional networks with independent parameters for hierarchical feature extraction, then produce two saliency region respectively; 3) The two saliency regions are merged by element-wise multiply; 4) The serial of merged saliency regions are resized and concatenated to produce the final saliency map.

\par We describe the details of the three adopted strategies in the following.

\begin{figure}[!t]
\centering
    \includegraphics[width=3.3in]{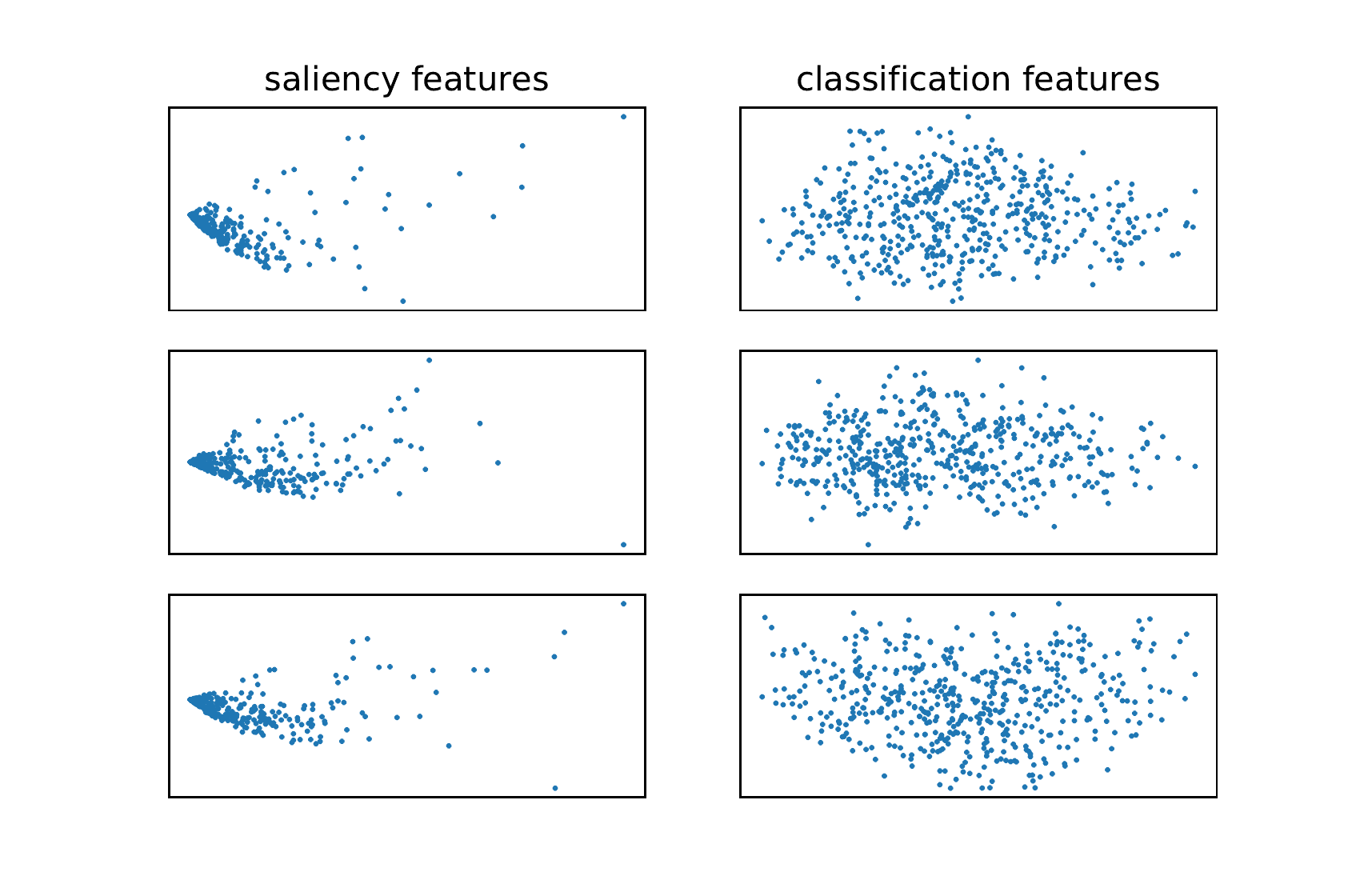}
    {PCA on VGG16 top layer feature maps from saliency detection and classification tasks.\label{fig1}}
\end{figure}

\subsection{Network Architecture}

\par \textbf{Depthwise Separable Convolution}: The fully convolutional feature extraction networks are build on stacked depth-wise separable convolutional blocks with inverted residual and linear bottleneck that introduced in MobileNetV2 \cite{b50}. Recent works with effort of designing more efficient neural networks are extensively adopting depth-wise separable convolution as the key building block, e.g. \cite{b47,b48,b49}. The basic idea for depth-wise separable convolution to achieve efficiency is separating the channel and spatial wise computation, which split one traditional $3\times3$ convolutional layer into one $3\times3$ depth-wise and one $1\times1$ point-wise convolutional layer.

\par Assume a convolutional layer with input channel of $C_{i}$, output channel of $C_{o}$, kernel size of $k$, and feature map size of $W\times H$, then the computation to produce the output feature map (assuming the input and output feature map size is consistent by zero padding) by traditional convolution is $Cost_{tra}=C_{i}\cdot C_{o}\cdot k\cdot k\cdot W\cdot\ H$. While the computation of depth-wise separable convolution with the same setting is $Cost_{dep}=C_{i}\cdot k\cdot k\cdot H\cdot W+C_{i}\cdot C_{o}\cdot 1\cdot 1\cdot H\cdot W$. Thus the proportion of computation between traditional and depth separable convolution can be represented as

\begin{equation}\frac{Cost_{tra}}{Cost_{dep}}=\frac{C_{o}\cdot k\cdot k}{C_{o}+k\cdot k}\end{equation}

\par Thus according to Equation.1, the bigger kernel size is, the more parameter get reduced. For a common choice of 128 output channel and $3\times3$ kernel, the depth-wise separable convolution reduces the parameter for approximately 8 times. Based on this, the adopted depth-wise separable convolutional block add one expansion layer with $1\times1$ kernel before the depth-wise convolution, which expand the input channels to support more sufficient feature extraction. As shown in Figure.2, the depth-wise separable convolutional block is consist of three layers of convolution: $1\times1$ expansion convolutional layer denote as $E^{t}$, $3\times3$ single channel convolution layer denote as $C^{t}$ and $1\times1$ bottleneck convolution layer denote as $B^{t}$.

\par We also adopt linear bottleneck and inverted residual on the depth-wise separable convolutional block, which use linear activation instead of ReLU activation after the bottleneck convolution layer, and add an element-wise plus operation $Out^{t}=B^{t}+B^{t-1}$ if the channel number is consistent. By stacking the bottleneck block and various convolutional layers, we construct the feature extraction network for our saliency detection approach. The detail settings of the network is shown in Table.1.

\begin{figure}[!t]
\centering
    \includegraphics[width=3.3in]{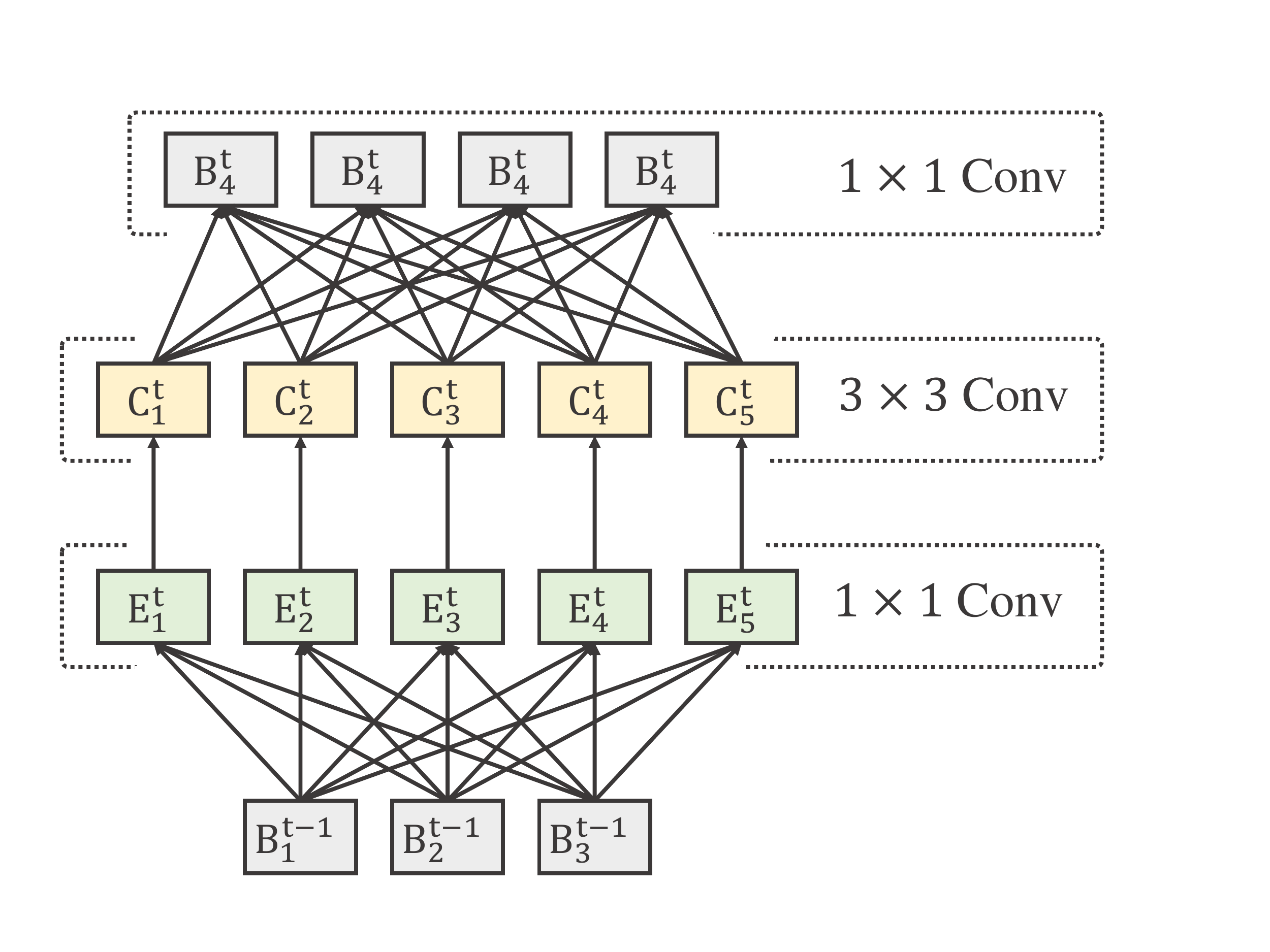}
    {Depth-wise separable convolutional block.\label{fig1}}
\end{figure}

\par \textbf{Regional Input}: To reduce the memory cost for temporarily storing feature maps with large spatial sizes, we replace the processing of the whole image to processing of serial of regions. For example, with the sample network structure, the memory cost for processing a $640\times480$ sized image is 48 times larger than processing a $80\times80$ sized image. Thus we crop the input image into regions and sequentially produce the saliency regions, then concatenate the output regions to the final saliency map.

\par Since we process sample images with cropped non-overlapping regions, the features are extracted from more local and object-oriented perspectives, resulting in limited representative power from more global perspective. Thus we apply multiscale feature extraction to learn richer semantics, with two networks to extract features from relatively coarse and fine resolution simultaneously. The two networks share the same structure shown in Table.1, and trained independently with fine and coarse resolution regions with size of $80\times80$. We obtain the two input regions following:

\begin{itemize}
\item Resizing the input image to short axis of 480 pixels and long axis accordingly to get fine input image, then copying the border to pad the fine input image by 80 at each end of axis to get coarse input image;
\item Cropping the fine resolution input image to serial of non-overlapping $80\times80$ sized regions;
\item Cropping a serial of $240\times240$ sized regions from coarse resolution input image at the same center location with fine resolution regions, then resizing to $80\times80$.
\end{itemize}

\par \textbf{Simplified Networks}: Before designing the network structure, we visualized the feature activation on the top layers from image classification methods and saliency detection methods that transferred from image classification networks. We find the activation of saliency detection is more densely activated and multiple nearly identical feature maps exist, while the activation of image classification method is sparse and no obviously identical activated feature maps exist. This indicates that when using the same structure, the classification method network is redundant for saliency detection task.

\par To avoid extra computational complexity caused by the redundancy, we build our network from scratch instead of transferring from exist model such as VGGNet \cite{b11}. The network includes 1 $3\times3$ kernel layer and 12 depth-wise separable convolutional block layers for feature extraction, with less channels at the bottom layers.

\begin{table}\centering
\caption{Feature extraction network structure}
\label{table}
\begin{tabular}{|c|c|c|c|}
\hline
Oprator & $S_{output}$ & $C_{expand}$ & $C_{output}$ \\
\hline
conv$3\times3$ & $80\times80$ & - & 32 \\
maxpool & $40\times40$ & - & 32 \\
block$3\times3$ & $40\times40$ & 64 & 16\\
block$3\times3$ & $40\times40$ & 64 & 16\\
maxpool & $20\times20$ & - & 16 \\
block$3\times3$ & $20\times20$ & 96 & 24\\
block$3\times3$ & $20\times20$ & 96 & 24\\
block$3\times3$ & $20\times20$ & 96 & 24\\
maxpool & $10\times10$ & - & 24 \\
block$3\times3$ & $10\times10$ & 128 & 32\\
block$3\times3$ & $10\times10$ & 128 & 32\\
block$3\times3$ & $10\times10$ & 128 & 32\\
block$3\times3$ & $10\times10$ & 128 & 32\\
block$3\times3$ & $10\times10$ & 256 & 64\\
block$3\times3$ & $10\times10$ & 256 & 64\\
block$3\times3$ & $10\times10$ & 512 & 128\\
conv$1\times1$ & $10\times10$ & - & 1 \\
\hline
\end{tabular}
\end{table}

\subsection{Model Training}

\begin{table*}\centering
\caption{Evaluation on MIT300 Benchmark Dataset}
\label{table}
\begin{tabular}{|c|c|c|c|c|c|c|c|c|}
\hline
Models & AUC-Judd$\uparrow$ & SIM$\uparrow$ & EMD$\downarrow$ & AUC-Borji$\uparrow$ & sAUC$\uparrow$ & CC$\uparrow$ & NSS$\uparrow$ & KL$\downarrow$ \\
\hline
our approach & 0.84 & 0.57 & 2.77 & 0.79 & 0.70 & 0.64 & 1.81 & 0.72 \\
Deep Gaze*\cite{b18} & 0.88 & 0.46 & 3.98 & 0.86 & 0.72 & 0.52 & 1.29 & 0.96 \\
Deep Fix\cite{b6} & 0.87 & 0.67 & 2.04 & 0.80 & 0.71 & 0.78 & 2.26 & 0.63 \\
SALICON\cite{b7} & 0.87 & 0.60 & 2.62 & 0.85 & 0.74 & 0.74 & 2.12 & 0.54 \\
Deep Gaze\cite{b18}  & 0.84 & 0.43 & 4.52 & 0.83 & 0.77 & 0.45 & 1.16 & 1.04 \\
PDP\cite{b21} & 0.85 & 0.60 & 2.58 & 0.80 & 0.73 & 0.70 & 2.05 & 0.92 \\
ML-Net\cite{b55} & 0.85 & 0.59 & 2.63 & 0.75 & 0.70 & 0.67 & 2.05 & 1.10 \\
BMS\cite{b37} & 0.83 & 0.51 & 3.35 & 0.82 & 0.65 & 0.55 & 1.41 & 0.81 \\
eDN\cite{b15} & 0.82 & 0.41 & 4.56 & 0.81 & 0.62 & 0.45 & 1.14 & 1.14 \\
Mr-CNN\cite{b38} & 0.79 & 0.48 & 3.71 & 0.75 & 0.69 & 0.48 & 1.37 & 1.08 \\
GBVS\cite{b39} & 0.81 & 0.48 & 3.51 & 0.80 & 0.63 & 0.48 & 1.24 & 0.87 \\
Judd Model\cite{b23} & 0.81 & 0.42 & 4.45 & 0.80 & 0.60 & 0.47 & 1.18 & 1.12 \\
LDS\cite{b40} & 0.81 & 0.52 & 3.06 & 0.76 & 0.60 & 0.52 & 1.36 & 1.05 \\
CAs\cite{b41} & 0.74 & 0.43 & 4.46 & 0.73 & 0.65 & 0.36 & 0.95 & 1.06 \\
\hline
\end{tabular}
\end{table*}

\begin{table*}\centering
\caption{Evaluation on CAT2000 Benchmark Dataset}
\label{table}
\begin{tabular}{|c|c|c|c|c|c|c|c|c|}
\hline
Models & AUC-Judd$\uparrow$ & SIM$\uparrow$ & EMD$\downarrow$ & AUC-Borji$\uparrow$ & sAUC$\uparrow$ & CC$\uparrow$ & NSS$\uparrow$ & KL$\downarrow$ \\
\hline
our approach & 0.85 & 0.64 & 1.66 & 0.76 & 0.57 & 0.73 & 1.93 & 0.96 \\
SAM\cite{b42} & 0.88 & 0.77 & 1.06 & 0.80 & 0.58 & 0.90 & 2.40 & 0.69 \\
DeepFix\cite{b6} & 0.87 & 0.74 & 1.15 & 0.81 & 0.58 & 0.87 & 2.28 & 0.37 \\
eDN\cite{b15} & 0.85 & 0.52 & 2.64 & 0.84 & 0.55 & 0.54 & 1.30 & 0.97 \\
BMS\cite{b37} & 0.85 & 0.61 & 1.95 & 0.84 & 0.59 & 0.67 & 1.67 & 0.83 \\
Judd Model\cite{b23} & 0.84 & 0.46 & 3.60 & 0.84 & 0.56 & 0.54 & 1.30 & 0.94 \\
LDS\cite{b40} & 0.83 & 0.58 & 2.09 & 0.79 & 0.56 & 0.62 & 1.54 & 0.79 \\
GBVS\cite{b39} & 0.80 & 0.51 & 2.99 & 0.79 & 0.58 & 0.50 & 1.23 & 0.80 \\
CAs\cite{b41} & 0.77 & 0.50 & 3.09 & 0.76 & 0.60 & 0.42 & 1.07 & 1.04 \\
\hline
\end{tabular}
\end{table*}

\par After the model construction, we train our approach using gradient descent. We use mean absolute error as loss function to calculate the distance between the output saliency region and label region, and use Adam optimizer \cite{b30} to update the parameter with the initial learning rate set to 0.001.

\par We train our approach on SALICON saliency detection dataset \cite{b19}. SALICON dataset is currently the biggest dataset for saliency detection task, with 10000 training samples, 5000 validation samples and 5000 testing samples from MS COCO dataset \cite{b29}. The stimuli sample set is consist of various indoor and outdoor scenes and objects with rich semantics. The ground truth fixation information is obtained by recording the mouse trajectory when multiple observers using mouse to direct their fixation on image stimuli during 5 seconds free viewing and 2 seconds followed waiting interval. To use SALICON to train our approach, each $640\times480$ sized sample image is cropped to 48 $80\times80\times2$ fine and coarse resolution regions, forming a new training set with 480000 samples.

\par At training phase, we use batch learning to accelerate convergence with batch size set to 48. The entire training takes about 24 hours on a 12GB RAM NVIDIA Tesla K40m GPU with the MxNet deep learning framework \cite{b31} on Ubuntu 16.04 operation system.

\section{Experimental Result}

\par In this section, we run experiments on both performance and computational cost evaluation. The performance result is to give a description on saliency detection performance of our approach. The computational cost result is to evaluate how much we achieve the light-weighted objective. We evaluate the performance on two benchmark datasets.

\par \textbf{MIT300} We mainly evaluate our model on the testing set of MIT300 \cite{b36} benchmark dataset. The MIT300 benchmark dataset is composed of 300 samples with various indoor and outdoor scenes and objects. The fixation information is extracted by directly recording the eye movements of 39 observers at 3 seconds free viewing at given sample. To avoid overfitting the dataset, the ground truth fixation maps are held out at the benchmark server for evaluation remotely, and the maximum submission is limited to 2 times per month. The sample sizes from MIT300 are ranged with x-axis from 679 to 1024 and y-axis from 457 to 1024, which are larger than samples from SALICON that we train our model on. Thus when evaluating on MIT300 dataset, we first resize the sample with short axis to 480 and long axis accordingly. The evaluation results are show in Table.2.

\par \textbf{CAT2000} We also evaluate our approach on CAT2000 \cite{b24} benchmark dataset. The CAT2000 dataset consists of one training set with accessable ground truth and one testing set with held out ground truth fixation maps. The training and testing set contains 20 different categories (100 images for each one) from \emph{Action} to \emph{Line Drawing}. The fixations are integrated from 5 seconds free viewing of 24 observers. Since the sample size of CAT2000 dataset is 1920$\times$1080, we resize the samples to 854$\times$480 for evaluation.

\subsection{Performance Metrics and Result}

\par At performance evaluation, multiple metrics are used, since previous study by Riche et al. \cite{b32} shows that no single metric has concrete guarantee of fair comparison. We briefly describe the used metrics for better understanding of the results. We denote $S$ for output saliency map, $G_{b}$ for ground truth fixation map with Gaussian blur and $G_{p}$ for ground truth fixation pixel map at following description.

\par \textbf{AUC}: Area Under ROC Curve (AUC) measures the area under the Receiver Operating Characteristic (ROC) curve, which consists of true and false positive rate under different binary classifier threshold between $S$ and $G_{p}$. Three different AUC implementations are mainly used in saliency detection, namely AUC-Judd \cite{b32}, AUC-Borji \cite{b33} and shuffled-AUC \cite{b34}. They are differed in how the true and false positive rate are calculated. The higher the true positive rate and the lower the false positive rate are, the larger the AUC is, and thus the better performance we have.

\par \textbf{EMD}: Earth Mover's Distance (EMD) normalizes $S$ and $G_{b}$ to two 2-dimensional distribution, and calculate the minimal cost of transferring $S^{N}$ to $G_{b}^{N}$. Thus lesser EMD score represents better performance.

\par \textbf{NSS}: Normalized Scanpath Saliency \cite{b35} is the mean value at on the fixation pixels location in normalized $S$ with zero mean and unit standard deviation. Larger NSS score represents better performance.

\begin{table*}\centering
\caption{Computational cost}
\label{table}
\begin{tabular}{|c|c|c|c|c|c|c|}
\hline
model & speed & total memory & net memory & model size & parameter & computation \\
\hline
unit & second per sample & MB & MB & MB & - & - \\
\hline
baseline: empty & 7.24$\times10^{-4}$ & \ \ \ \ 64.64 & 0 & 0 & 0 & 0 \\
our approach & 1.60656 & \ \ \ \  89.09 & \ \ \ \  24.45 & \ \ \ \  0.94 & \ \ \ \  245,440 & \ \ \ \ 3,577,036,800 \\
SALICON\cite{b7} & 4.15158 & 1275.17 & 1210.53 & 112.41 & 14,711,488 & 187,094,630,400 \\
Deepfix\cite{b6} & 3.46976 & 1248.95 & 1184.31 & 110.02 & 28,805,312 & 117,449,932,800 \\
Deepgaze\cite{b18} & 6.18503 & 2502.68 & 2438.04 & 76.64 & 20,060,418 & 119,489,205,600 \\
SalGAN\cite{b54} & 9.65656 & 2033.69 & 1969.05 & 121.40 & 29,456,105 & 198,711,754,800 \\
PDP\cite{b21} & 3.66918 & 1091.23 & 1026.59 & 59.32 & 15,526,216 & 96,825,302,400 \\
ML-Net\cite{b55} & 5.40197 & 1104.33 & 1039.69 & 59 & 15,447,808 & 124,866,048,000 \\
\hline
\end{tabular}
\label{tab3}
\end{table*}

\par \textbf{CC}: Correlation Coefficient (CC) measures the linear relationship between saliency matrix $S$ and $G_{b}$. CC score of 1 means $S$ and $G_{b}$ are identical, while 0 means $S$ and $G_{b}$ are uncorrelated. Thus larger CC score represents better performance.

\par \textbf{Sim}: Similarity (Sim) first normalizes $S$ and $G_{b}$ to $S^{N}$ and $G^{N}_{b}$, then calculate the sum of element-wised minimum between $S^{N}$ and $G^{N}_{b}$. Thus larger Similarity represent better performance.

\par \textbf{KLD}: Kullback-Leibler Divergence (KLD) is a non-symmetric metric. It measures the information lost when using $S$ to encode $G_{b}$. Lesser KLD score represents better saliency detection performance.

\par We evaluate the saliency detection performance of our approach on test set of MIT300 and CAT2000 benchmark datasets, the results are shown in Table.2 and Table.3 respectively. From the tables we can see that our approach achieves competitive results comparing with previous methods transferred from large scale image classification models, and outperforms traditional and shallow network based methods.

\subsection{Computational Cost Measurements and Result}

\par As the goal of this work is to propose a light-weighted DNNs based architecture for saliency prediction, we evaluate the computational complexity of our approach and some previous works. We briefly describe the measurements we used for computational complexity evaluation. We run all the tests on a Intel Core i7-4710MQ CPU with MxNet deep learning framework.

\par \textbf{Speed}: speed is the time cost of processing one $640\times480$ sized image. Since our approach process $640\times480$ sized image by processing the cropped 48 $640\times480$ sized region, we evaluate the speed of our approach by measuring the average time cost of processing a whole image with the cropping and concatenating procedure.

\par \textbf{Memory Cost}: memory cost is the average RAM consumption when processing one $640\times480$ sized image. Since the deep learning framework have extra memory requirement besides the actual model consumed RAM, we give two results for memory cost: one for total memory cost when running the model, one for the net consumption which is the rough cost minus the framework minimal RAM requirement.

\par \textbf{Model Size}: we evaluate the model size by measuring the bytes of the parameter file for each model. The parameter file is a cross platform .params file.

\par \textbf{Parameter Size}: parameter size is the total number of learnable parameters for each model such as the convolution kernels and fully connection weights. The parameter is calculated follows the equation $$p=\sum_{l=1}^{L}C_{l-1}\cdot C_{l}\cdot k\cdot k$$ where $k$ denotes kernel size and $C_{l}$ denotes the channel number of layer $l$. Notice that $C_{0}$ denotes the input channel, for RGB image $C_{0}=3$.

\par \textbf{Computation Size}: computation size is the total number of computation for processing one $640\times480$ sized image. For our approach, we measure the total computation by multiply the computation number of processing two $80\times80$ regions (coarse and fine resolution regions) by 48. The computation is calculated follows the equation $$p=\sum_{l=1}^{L}C_{l-1}\cdot C_{l}\cdot k\cdot k\cdot W\cdot H$$ where $W$ and $H$ denotes the feature map width and height.

\par The results are shown in Table.4. We compared the computational cost of our approach with some deep learning based previous works on the described measurements. Notice that when running neural networks, the framework (in our case MxNet) has a minimum resource consumption. Thus we run a emtpy network with no operation as baseline for minimum computational cost and further calculate the actual computational cost of the networks.

\par We can see that the net memory cost of our approach is 42 (PDP\cite{b21}) to 99 (DeepGazeII\cite{b18}) times less than the previous works, the model size is 63 (ML-Net\cite{b55}) to 129 (SalGAN\cite{b54}) times smaller. The parameter size of our approach is 63 (ML-Net) to 117 (DeepFix\cite{b6}) times smaller, and the computation size is 27 (PDP) to 56 (SalGAN) times smaller. While the computational cost is distinctively reduced, the time cost for processing one $640\times480$ sized image is still less than other previous methods.

\section{Conclusion}

\par In this work, we propose a light-weighted neural networks architecture for saliency detection with distinctively lower memory cost and model size, while maintaining competitive performance comparing with previous approaches. We mainly adopt three strategies to achieve light-weight goal: 1) depth-wise separable convolutional block to reduce parameter number; 2) regional input to reduce intermediate feature map size; 3) simplified network structure to reduce model size. Experimental result shows that our approach reduce the runtime memory cost by 42 to 99 times, and model storage size by 63 to 129 times comparing with previous approaches.


\begin{thebibliography}{00}

\bibitem{b1} Marchesotti, Luca, C. Cifarelli, and G. Csurka. "A framework for visual saliency detection with applications to image thumbnailing." IEEE, International Conference on Computer Vision IEEE, 2009:2232-2239.

\bibitem{b2} Achanta, Radhakrishna, and S. Susstrunk. "Saliency detection for content-aware image resizing." IEEE International Conference on Image Processing 2009:1005-1008.

\bibitem{b3} Borji, Ali, et al. "Online learning of task-driven object-based visual attention control." Image and Vision Computing 28.7 (2010): 1130-1145.

\bibitem{b4} Dankers, Andrew, N. Barnes, and A. Zelinsky. "A Reactive Vision System: Active-Dynamic Saliency." 2007.

\bibitem{b6} Sss, Kruthiventi, K. Ayush, and R. V. Babu. "DeepFix: A Fully Convolutional Neural Network for Predicting Human Eye Fixations." IEEE Transactions on Image Processing 26.9(2017):4446-4456.

\bibitem{b7} Huang, Xun, et al. "SALICON: Reducing the Semantic Gap in Saliency Prediction by Adapting Deep Neural Networks." IEEE International Conference on Computer Vision IEEE Computer Society, 2015:262-270.

\bibitem{b8} Treisman, Anne, and Garry Gelade. "A feature-integration theory of attention." Cognitive Psychology 12.1 (1980): 97-136.

\bibitem{b9} Koch, C, and S. Ullman. "Shifts in selective visual attention: towards the underlying neural circuitry." Hum Neurobiol 4.4(1987):219-227.

\bibitem{b10} Itti, Laurent, C. Koch, and E. Niebur. A Model of Saliency-Based Visual Attention for Rapid Scene Analysis. IEEE Computer Society, 1998.

\bibitem{b11} Simonyan, Karen, and A. Zisserman. "Very Deep Convolutional Networks for Large-Scale Image Recognition." Computer Science (2014).

\bibitem{b15} Vig, Eleonora, M. Dorr, and D. Cox. "Large-Scale Optimization of Hierarchical Features for Saliency Prediction in Natural Images." Computer Vision and Pattern Recognition IEEE, 2014:2798-2805.

\bibitem{b16} Kummerer, Matthias, L. Theis, and M. Bethge. "Deep Gaze I: Boosting Saliency Prediction with Feature Maps Trained on ImageNet." Computer Science (2014).

\bibitem{b17} Krizhevsky, Alex, I. Sutskever, and G. E. Hinton. "ImageNet classification with deep convolutional neural networks." International Conference on Neural Information Processing Systems Curran Associates Inc. 2012:1097-1105.

\bibitem{b18} Kummerer, Matthias, T. S. A. Wallis, and M. Bethge. "DeepGaze II: Reading fixations from deep features trained on object recognition." (2016).

\bibitem{b19} Jiang, Ming, et al. "SALICON: Saliency in Context." Computer Vision and Pattern Recognition IEEE, 2015:1072-1080.

\bibitem{b20} Szegedy, Christian, et al. "Going deeper with convolutions." IEEE Conference on Computer Vision and Pattern Recognition IEEE Computer Society, 2015:1-9.

\bibitem{b21} Jetley, Saumya, N. Murray, and E. Vig. "End-to-End Saliency Mapping via Probability Distribution Prediction." Computer Vision and Pattern Recognition IEEE, 2016:5753-5761.

\bibitem{b23} Judd, T, et al. "Learning to predict where humans look." IEEE, International Conference on Computer Vision IEEE, 2010:2106-2113.

\bibitem{b24} Borji, Ali, and L. Itti. "CAT2000: A Large Scale Fixation Dataset for Boosting Saliency Research." Computer Science (2015).

\bibitem{b29} Lin, Tsung Yi, et al. "Microsoft COCO: Common Objects in Context." 8693(2014):740-755.

\bibitem{b30} Kingma, Diederik, and J. Ba. "Adam: A Method for Stochastic Optimization." Computer Science (2014).

\bibitem{b31} Chen, Tianqi, et al. "MXNet: A Flexible and Efficient Machine Learning Library for Heterogeneous Distributed Systems." Statistics (2015).

\bibitem{b32} Riche, Nicolas, et al. "Saliency and Human Fixations: State-of-the-Art and Study of Comparison Metrics." IEEE International Conference on Computer Vision IEEE, 2014:1153-1160.

\bibitem{b33} Borji, Ali, et al. "Analysis of Scores, Datasets, and Models in Visual Saliency Prediction." IEEE International Conference on Computer Vision IEEE Computer Society, 2013:921-928.

\bibitem{b34} Zhang, L., et al. "SUN: A Bayesian framework for saliency using natural statistics. " J Vis 8.7(2008):32.1.

\bibitem{b35} Peters, R. J., et al. "Components of bottom-up gaze allocation in natural images." Vision Research 45.18(2005):2397-2416.

\bibitem{b36} Judd, Tilke, F. Durand, and A. Torralba. "A Benchmark of Computational Models of Saliency to Predict Human Fixations." (2012).

\bibitem{b37} Zhang, Jianming, and S. Sclaroff. "Saliency Detection: A Boolean Map Approach." IEEE International Conference on Computer Vision IEEE Computer Society, 2013:153-160.

\bibitem{b38} Liu, Nian, et al. "Predicting eye fixations using convolutional neural networks." Computer Vision and Pattern Recognition IEEE, 2015:362-370.

\bibitem{b39} Sch?lkopf, Bernhard, J. Platt, and T. Hofmann. "Graph-Based Visual Saliency." International Conference on Neural Information Processing Systems MIT Press, 2006:545-552.

\bibitem{b40} Fang, Shu, et al. "Learning Discriminative Subspaces on Random Contrasts for Image Saliency Analysis." IEEE Transactions on Neural Networks \& Learning Systems 28.5(2017):1095-1108.

\bibitem{b41} Goferman, S, L. Zelnikmanor, and A. Tal. "Context-aware saliency detection. " IEEE Transactions on Pattern Analysis \& Machine Intelligence 34.10(2012):1915-1926.

\bibitem{b42} Cornia, Marcella, et al. "Predicting Human Eye Fixations via an LSTM-based Saliency Attentive Model." (2016).

\bibitem{b47} Zhang, Xiangyu, et al. "ShuffleNet: An Extremely Efficient Convolutional Neural Network for Mobile Devices." (2017).

\bibitem{b48} Howard, Andrew G, et al. "MobileNets: Efficient Convolutional Neural Networks for Mobile Vision Applications." (2017).

\bibitem{b49} Chollet, Francois. "Xception: Deep Learning with Depthwise Separable Convolutions." (2016):1800-1807.

\bibitem{b50} Sandler, Mark, et al. "MobileNetV2: Inverted Residuals and Linear Bottlenecks." (2018).

\bibitem{b51} Cong, Jason, and B. Xiao. Minimizing Computation in Convolutional Neural Networks. Artificial Neural Networks and Machine Learning - ICANN 2014. Springer International Publishing, 2014:281-290.

\bibitem{b52} Iandola, Forrest N, et al. "SqueezeNet: AlexNet-level accuracy with 50x fewer parameters and <0.5MB model size." (2016).

\bibitem{b53} Han, Song, H. Mao, and W. J. Dally. "Deep Compression: Compressing Deep Neural Networks with Pruning, Trained Quantization and Huffman Coding." Fiber 56.4(2015):3--7.

\bibitem{b54} Pan, Junting, et al. "SalGAN: Visual Saliency Prediction with Generative Adversarial Networks." (2017).

\bibitem{b55} Cornia, Marcella, et al. "A deep multi-level network for saliency prediction." international conference on pattern recognition (2016): 3488-3493.

\end{thebibliography}
\end{document}